\documentclass{article}

\usepackage{PRIMEarxiv}

\usepackage[utf8]{inputenc} 
\usepackage[T1]{fontenc}    
\usepackage{hyperref}       
\usepackage{url}            
\usepackage{booktabs}       
\usepackage{amsfonts}       
\usepackage{nicefrac}       
\usepackage{microtype}      
\usepackage{lipsum}
\usepackage{fancyhdr}       
\usepackage{graphicx}       
\usepackage{subfigure}
\usepackage{amsmath}
\usepackage{algorithm}
\usepackage{multirow}
\usepackage{algorithmic}
\graphicspath{{media/}}     
\usepackage{natbib}
\setcitestyle{authoryear,round,citesep={;},aysep={,},yysep={;}}

\title{QuickNets: Saving Training and Preventing Overconfidence in Early-Exit Neural Architectures
}

\author{
  Devdhar Patel, Hava Siegelmann \\
  Biologically Inspired Neural and Dynamical Systems Laboratory (BINDS) \\
  College of Computer and Information Sciences, \\
University of Massachusetts Amherst \\
   Amherst, MA 01003, USA\\
  \texttt{\{devdharpatel, hava\}@cs.umass.edu} \\
}

\begin{document}
\maketitle

\begin{abstract}
Deep neural networks have long training and processing times. Early exits added to neural networks allow the network to make early predictions using intermediate activations in the network in time-sensitive applications. However, early exits increase the training time of the neural networks. We introduce QuickNets: a novel cascaded training algorithm for faster training of neural networks. QuickNets are trained in a layer-wise manner such that each successive layer is only trained on samples that could not be correctly classified by the previous layers. We demonstrate that QuickNets can dynamically distribute learning and have a reduced training cost and inference cost compared to standard Backpropagation. Additionally, we introduce commitment layers that significantly improve the early exits by identifying for over-confident predictions and demonstrate its success.
\end{abstract}


\section{Introduction}
\label{introduction}
Deep networks have made a substantial progress on many challenging machine learning tasks. To achieve this, state-of-the-art networks have become larger, deeper and more complex over the years. As a consequence, the demand on computing resources, including computation time, required for training and inference of deep networks have grown over the years \cite{he2016deep, krizhevsky2012imagenet}. However, a vast majority of real-world applications for deep networks have constraints on resources and time. It is an open research challenge to achieve similar results in a constrained setting. With the emergence of 5G and Fog Computing, this problem is gaining more attention than ever.

To reduce the footprint of deep networks, various methods for lowering complexity were suggested by lowering the computational precision of the network using methods like compression and quantization have been proposed \cite{Han2016DeepCC}. These methods are based on the trade-off between precision and computational time and thus perform worse than larger networks. In contrast, multi-output networks, branching networks and early-exit networks add more layers to the networks allowing for faster inference at the cost of network size. Early-exit networks add multiple early-exits to the neural network so that the entire network is not used for every single input. Early-exit networks are based on the principle that not all inputs to a network have the same complexity and thus not all inputs require the same processing time during inference \cite{Ma2020WhyLL}. Therefore, early-exit networks allow majority of input patterns to exit early while reserving the layer exits for more complex inputs. 


State of the art early-exit networks are trained either in an end-to-end fashion, training all the exits simultaneously using a combined loss \citep{Zhou2019ElasticNN} or the backbone (all layers required to train the last exit) is trained first and the early-exits are then attached and trained while freezing the weights of the backbone \citep{Panda2016ConditionalDL, Teerapittayanon2016BranchyNetFI, Bakhtiarnia2021ImprovingTA}. While these approaches allow for faster inference, they require an even longer training time than the original network in order to achieve this. 

A multi-exit network can be distributed across different hardware. For instance, a battery powered robot on field and a control center. Since the robot is energy constrained and it might not be feasible to run a large neural network on it. On the other hand, due to communication latency and privacy, running the entire network at the control center might also not be desirable. Early exit networks provide a way of distributing the compute between the two so that the robot can make quick decisions for simple inputs while deferring the difficult to the larger network at the control center. In this setting, it is important to also consider the network training in a distributed manner. Our QuickNets will provide a solution towards this end.

To build a truly modular and computationally adaptive network, we need to address all the problems described above. The core idea of early-exits is to provide an early inference for the easy input samples while reserving the later layers for more complex inputs. However, state of the art existing approaches do not take this into account while training. For our approach, we take inspiration from neuroscience. The brain has connections to the superior colliculus from almost every other area \citep{Harting1992CorticotectalPI}. The superior colliculus is a site of sensorimotor integration and thus can be thought of as the "output" of the brain. This architecture might be related to the adaptive response rate of the brain. Additionally, it has also been shown that the brain has depth-wise hierarchical organization with increasing function abstraction \cite{Taylor2015TheGL}. This suggests that the deeper areas layers in the brain are trained on the harder and more abstract tasks while simpler tasks use, and thus trained on, fewer layers.

\begin{figure*}[ht]
\begin{center}
\centerline{\includegraphics{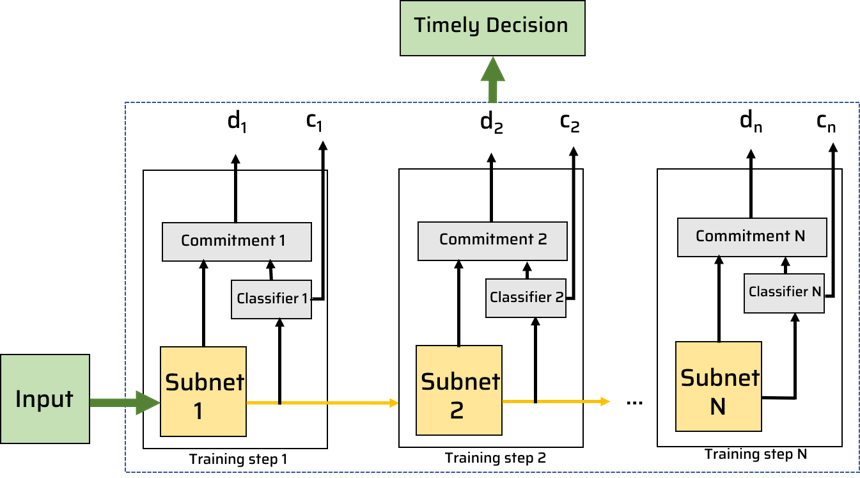}}
\caption{High level demonstration of the Quick-witted neural training algorithm. At each training step, a sub-network is trained on inputs that could not be solved by the previous sub-networks. Note that the error does not need to back propagate between different sub-networks. A fast inference can be made for inputs that can be solved early without computing the deeper layers. The architecture is adaptive and deeper layers can be iteratively added to solve harder inputs.}
\label{fig:network}
\end{center}
\end{figure*}

In this work, we introduce a brain inspired Quick-Witted Neural Network (QuickNet), that demonstrates a similar ability to distribute learning across layers depending on the difficulty of the task. This novel network (see Figure \ref{fig:network}) is trained in a block-wise manner and has outputs at each block for early inference of trivial inputs while more complex inputs may be processed in the deeper layers. We introduce a novel training algorithm that trains the network one exit at a time so that each exit is only trains layers following the last exit on samples that could not be correctly classified by the previous layers. In order to achieve this, we also introduce our "commitment layers" for each exit that identify overconfident predictions, thus making the decision to exit more reliable. 

We demonstrate that our approach on a variety of  image classification datasets 
and show it reduces the training cost to less than half when compared to end-to-end trained backpropagation networks with minimal reduction of performance. The cost of inference is also reduced significantly. Additionally, Since this approach trains blocks of the network at a time, the memory requirement while training is drastically reduced as opposed to end-to-end training which requires intermediate partial gradients to be saved for all the layers in the network. 
We believe that our work is a significant step towards distributed hierarchical learning in resource constrained setting. 

\section{Related work}

\textbf{Early-exit networks}: Early exit networks have demonstrated that inference time on large networks can be reduced significantly by adding early exits to pre-trained networks. \citet{Panda2016ConditionalDL} showed that adding a single linear classifier at early layers of deep networks significantly reduces the average operations per input. \citet{Teerapittayanon2016BranchyNetFI} on the other hand, added branches to well-known deep networks like AlexNET, LeNET and ResNET and showed speed up betweem 1.9x to 5.4x. Each branch consists of one or more convolution layers followed by a classifier Both these approaches require manual selection of place to add the exits / branches. \citet{Huang2018MultiScaleDN} avoided this problem by creating dense networks such that each layer is connected to every exit. Their network is also designed such that every exit has access to coarse and fine features. Generally, earlier convolution layers only have access to fine features. However, their approach is designed for very specific network architecture and cannot be generalized to all neural network architectures. 

Another problem with early-exit network is that earlier layers have a large number of features which reduce the accuracy of early exits. Generally, early exit networks address by adding max or average pooling layers. However, these can result in loss of information. \citet{Passalis2020EfficientAI} addressed this problem by a using bag-of-feature pooling and showed that it improved the accuracy over traditional pooling. This problem is beyond the scope of this work, however, our approach can easily be used with bag-of-features pooling. 

Recent works on early-exits have also focused on using early exits on problems other than image classification. \citet{Zhou2020BERTLP} proposed an early-exit method that can be used on language models. \citet{Ghodrati2021FrameExitCE} introduced an early-exit framework for video recognition that outperforms state-of-the-art efficient video recognition methods.

Finally, another area of research in early-exit networks is the better training of early-exits. \citet{Li2019ImprovedTF} introduced three techniques to better train early-exits including gradient re-scaling, using earlier predictions as a prior for later exits, and knowledge distillation using the final exit as a teacher. However, these techniques can only be applied to end-to-end training. \citet{Bakhtiarnia2021ImprovingTA} uses curriculum learning improve the accuracy early-exits. However, in their approach, the do not tune the weights of the backbone network.

An alternative way of training an early-exit neural networks is to train all the exits together by minimizing a joint loss. \citet{Zhou2019ElasticNN} demonstrate that this can be an effective way of reducing the problem of vanishing gradient. However, they did not study their network in an early inference setting. 

Orthogonal to adding early-exits to network, another line of work focuses on efficient inference using adaptive paths inside the neural network. The popular ResNet architecture has been shown to function as an ensemble of shallow networks \citep{Veit2016ResidualNB}.  Adaptive inference methods exploit this phenomenon to gate blocks of networks for more efficient inference. \citep{Veit2018ConvolutionalNW, Wang2018SkipNetLD, Wu2018BlockDropDI}. Note that early exit networks are related to this approach. The early-exits can be viewed as a skip connection to the final layer of the network.

Our approach differs from all other approaches as it focuses on reducing the training cost along with the inference cost.

\textbf{Exit criterion:} Early-exit networks need a criteria to make the decision of choosing an early exit. By far, the most popular criteria is the confidence of an exit \citep{Scardapane2020WhySW}. The confidence of an exit is measured by the entropy of the output as follows:

\begin{equation}
    entropy(y) = \frac{1}{\log(C)}\sum_{c \in C}y_c \log y_c
\end{equation}

Where $y$ is a vector containing the probabilities for all the class labels and $C$ is the set of all class labels. Thus, lower entropy corresponds to higher confidence. However, confidence of each exit may vary and the confidence threshold of each exit requires tuning according to its accuracy. Even after tuning, using confidence might lead to poor results since neural networks are often overly confident for incorrect predictions \cite{Vemuri2020ScoringCI}. Recently, several works have explored strategies to improve the confidence estimation. \citet{Enomoto2021LearningTC} added a confidence term to the training loss, thus calibrating the confidence while training. However, their method cannot be used in layer-wise training since the new loss term uses prediction of the last exit. 

Another approach for choosing an exit is to implement decision gates. \citet{Shafiee2019DynamicRT} implemented decision gates based on the distance of the sample from the decision boundaries. \citet{Scardapane2020DifferentiableBI} implemented differentiable gates that can be trained as a soft conditional output for each branch. Their approach also allows the inference to be based on all the previous output, weighted by their confidence. However, their method cannot be used in a layer-wise training. \citet{Wang2018IDKCF} introduced search based methods for creating an augmenting classifier which is similar to the decision layer in our approach. However, in our approach, instead of making the decision whether to exit, the commitment layers are used to identify overconfident predictions. The decision to exit is thus made using both the confidence and the commitment layers.

\textbf{Layer-wise training:} Layer-wise training has been a fundamental area of research in neural networks. In fact, some works on layer-wise training predate backpropagation. In the late 1980's, \citet{Fahlman1989TheCL} introduced an algorithm to sequentially train perceptrons. Recently, \citet{Marquez2018DeepCL} built on their work to introduce a cascaded layer-wise learning algorithm for deep neural networks. They found that layer-wise trained network had an inferior final accuracy to networks trained end-to-end using backpropagation. However, as compared to the end-to-end network, the cascaded network had features that were more correlated to the final output in earlier layers. They showed that fine-tuning the cascaded network using backpropagation can produce an network with better performance than the end-to-end trained network. 

Since then, there has been steady progress in the area of layer-wise training. \citet{Nkland2019TrainingNN} added an additional similarity matching loss to layer-wise training and showed that layer-wise training can approach state-of-the-art.
\citet{Belilovsky2019GreedyLL} showed that layer-wise training can scale to the ImageNet dataset. They show that training the network one layer at a time can outperform the ImageNet performance on AlexNet architecture \citep{krizhevsky2012imagenet}. They also demonstrate that training two and three layers at a time, their network can beat ImageNet performance on the VGG model family \citep{simonyan2014very}. However, their approach uses a large number of filters in order to achieve this result.

\citet{Belilovsky2020DecoupledGL} introduced an algorithm to train all the layers parallel while keeping the training layer-wise by stopping the gradient flow down the network from each exit. Their approach can be easily applied to our algorithm to speed up training and also train each layer asynchronously, however, we leave that to future work.

\citet{Ma2020WhyLL} studied why layer-wise training is harder than backpropagation. They argue that features from shallow layers are poorly separable and thus supervised learning in earlier layers should be weakened using aggressive down-sampling. \citet{Zhang2021TrainYC} on the other hand, argue that the bottoms up training approach of the layer-wise training leads to worse performance since features from earlier layers have poor transfer-ability across different networks for the same dataset. Instead, they show that training the network top-down from classifier can outperform the end-to-end network. However, their approach requires training the entire network in an end-to-end fashion and then iteratively freezing the layers from the end and re-initializing the preceding layers and training them again. This requires more training time than the end-to-end trained network and is not designed for earlier inference.

\citet{Mostafa2018DeepSL} Introduced a biologically plausible way of layer-wise training using local errors from fixed random auxiliary classifiers. \citet{Du2019TransferLA} Demonstrated that cascaded networks outperform end-to-end networks on transfer learning tasks. \citet{Skolik2021LayerwiseLF} Introduced layer-wise training for quantum neural networks. 

To our knowledge, ours is the first work to explore the reduction of the training dataset for as layer-wise training progresses to deeper layers of the network.

\section{Methods}

In this section, we formalize the architecture and training algorithm of the Quick-witted neural networks.  

\subsection{Network architecture}

The QuickNet network architecture has $N$ blocks (see Fig.\ref{fig:network}), trained in succession. Thus, the network requires $N$ steps to be trained. At each step, a part of the backbone of the entire network is trained. The subnet in each block is made up of one or more layers that belong to the backbone. The classifier in each block may also contain one or more layer that take the output of the subnet and output a prediction. The parameters of subnet for block $i$ are denoted $\theta_i$ and the parameters of the classifier are denoted by $\gamma_i$. Let $W_{\theta_i}$ be the learnable function parameterized by $\theta_i$, then for each block, we get:

\begin{align}
\nonumber& x_{i,j} = W_{\theta_i}x_{i-1,j}, \text{where } x_{i-1, j} \in T_{i-1} \\
\nonumber& c_{i} = W_{\gamma_i}x_i \\
\nonumber& d_{i} = W_{\delta_i}x_i \\
\nonumber& X_i = \{x_{i,1}, x_{i,2}, ... \} \\
\nonumber& L_i = \text{set of all learned outputs }\\
\nonumber& T_i = X_i - L_{i} \\
\end{align}

Where $x_{i-1,j}$ is the input to the block $i$ with index $j$. ${x_0}$ is the input image and $T_0$ is the set of all inputs. The classifier $c_i$ and subnet are trained using the cross-entropy loss of $c_i$. The commitment layer parameters $\delta_i$ are then trained on the binary-cross entropy loss using the correctness of the prediction $c_i$ as a label. At training step $i$, the parameters of the previous blocks are frozen and are not changed.

\subsection{Removing samples from training data}
The network is trained in a block-wise manner. First block is trained on all inputs. Once the a block is trained, learned training samples are removed from the dataset. An input is considered learned by a particular subnet if the following conditions are met:
\begin{enumerate}
    \item The confidence on the sample is above the threshold.
    \item The commitment layer output is positive. 
    \item The sample is correctly classified.
\end{enumerate}
In order to keep the dataset balanced after removing the learned samples, the sampling frequency during training is adujusted for each class proportional to the number of remaining samples.
Algorithm \ref{alg:example} describes the training process of QuickNets.
\begin{algorithm}[H]
   \caption{QuickNet training}
   \label{alg:example}
\begin{algorithmic}
   \STATE {\bfseries Input:} Training samples $T_0 = \{x_{0,n}, y_n\}_{n\leq N}$
   \FOR{$j \in 0..M-1$}
   
   \STATE Train Block $j$ parameters $W_{\theta_{j}}$  on samples $T_j$
   \STATE Train commitment parameters $W_{\delta_{j}}$ on samples $T_j$
   \STATE Get learned samples $L_j$
   \STATE $T_{j+1} = T_j - L_j$
  \ENDFOR
\end{algorithmic}
\end{algorithm}

\subsection{Inference}
The decision to pick and exit is made based on the confidence of the exit. The confidence of an exit is measured by the entropy of the output as follows:

\begin{equation}
    confidence(y) = \frac{1}{C}\sum_{c \in C}y_c \log y_c
\end{equation}

During inference, an early exit is chosen if the confidence value $d_i$ is greater than the threshold $t$ and the commitment layer output is positive. If the confidence values of all exits are below threshold value, the output of exit with the highest confidence value is chosen. Thus, the threshold value $t$ allows for the trade-off between accuracy and inference speed. A lower threshold value allows for faster inference while reducing the accuracy. Note that it is also common to choose the last exit in case none of the decision values are above the threshold \citep{Teerapittayanon2016BranchyNetFI, Scardapane2020WhySW}. Other choices include the use of all exits as an ensemble \cite{Belilovsky2019GreedyLL}. However, choosing the exit with the highest decision value allows us to choose the exit best suited for an input thus mitigating the overthinking problem \citep{Kaya2019ShallowDeepNU} which is more prominent in QuickNets since the deeper layers are only trained on a subset of the data.

\section{Experiments} 
In this section, we examine the performance of QuickNets and analyze its training cost, inference cost and behaviour. We also examine the dynamic reduction of training samples and the distribution of learning across blocks. Finally, we examine the efficacy of the commitment layers when used on existing early exit approaches.

In every experiment we train the networks/blocks until convergence or 100 epochs (whichever comes first). We use three datasets:  MNIST \citep{lecun1998gradient}, Fashion-MNIST \citep{xiao2017fashion} and CIFAR-10 \citep{krizhevsky2009learning}. For MNIST and Fashion MNIST, we use a fully connected architecture where each block contains a single layer with 150 neurons. CIFAR-10 is trained on the VGG architecture \cite{simonyan2014very} with each block containing one or more convolutional layers. The internal classifiers employ mixed max-average pooling strategy \cite{lee2016generalizing} for feature reduction. The training and hyperparameters are discussed in more detail in the Appendix.

To evaluate the computation cost of training and inference, we calculate the average floating-point operations required for perform a single forward pass to each exit. We use this as the hardware-agnostic measure of cost and time-cost of training and inference. 

\begin{figure*}[t]
\includegraphics[width=0.5\columnwidth]{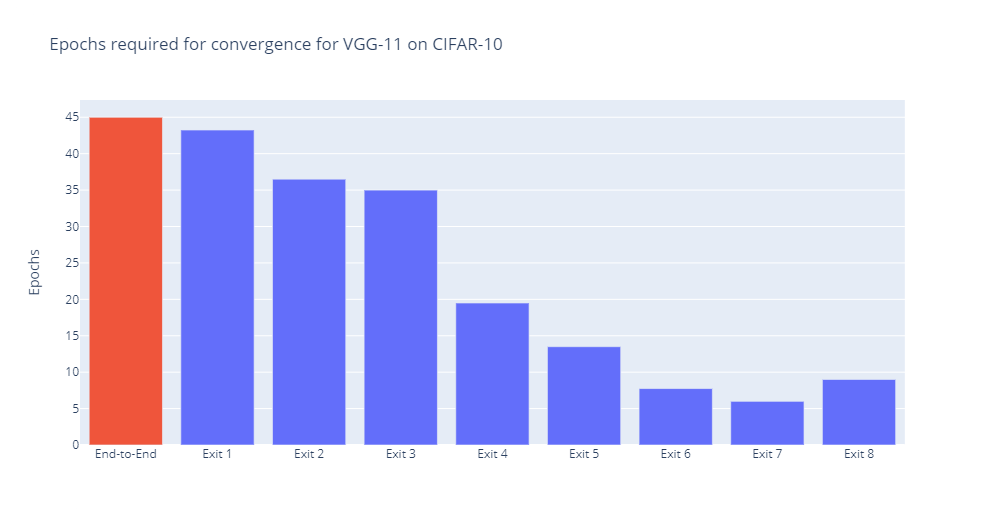}
\includegraphics[width=0.5\columnwidth ]{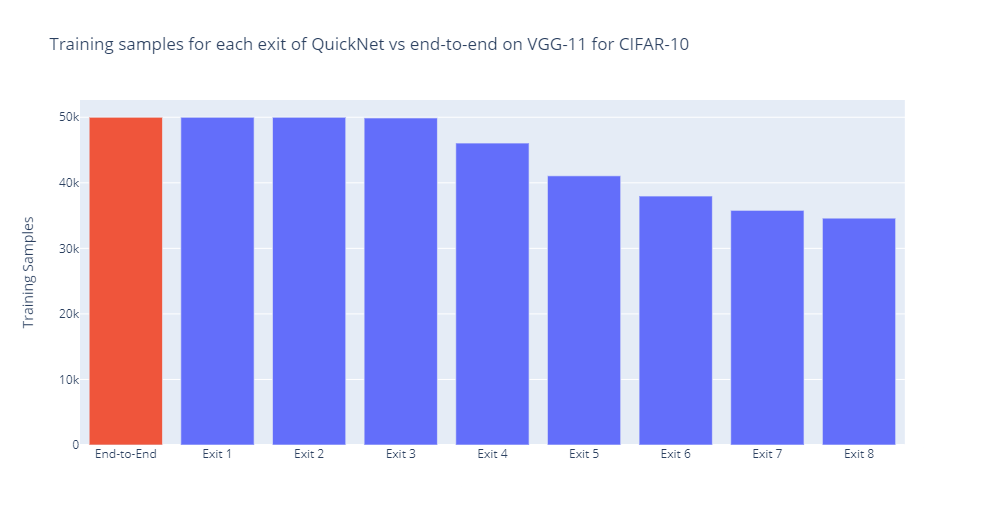}
\includegraphics[width=0.5\columnwidth ]{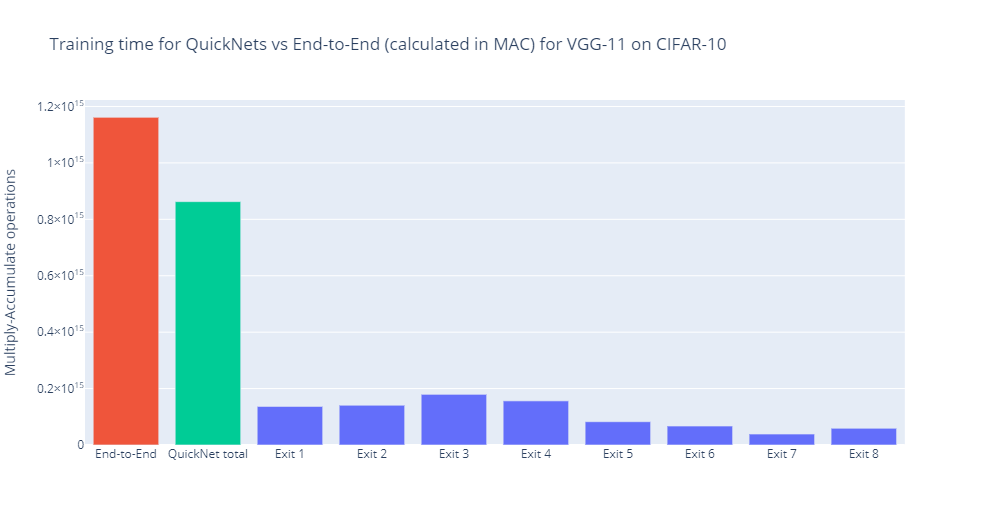}
\includegraphics[width=0.5\columnwidth ]{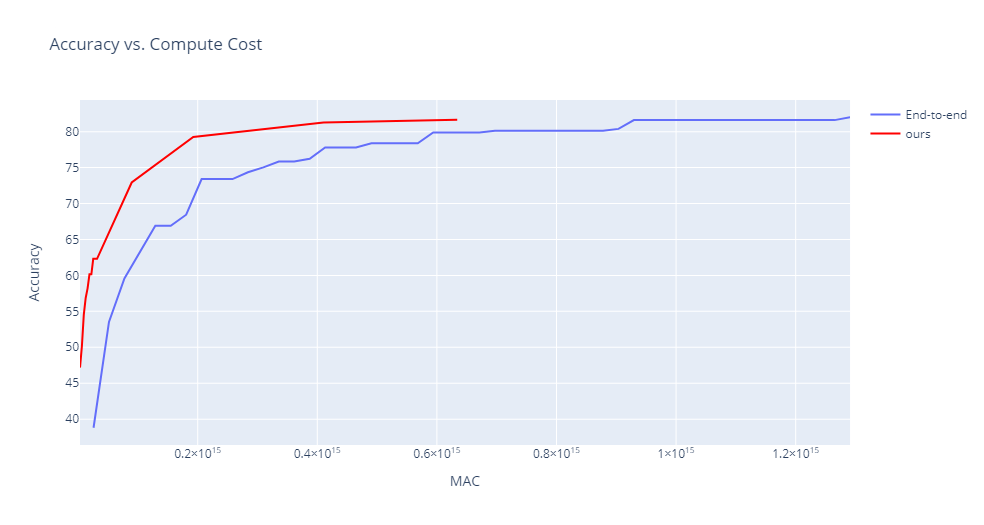}
\caption{Top-left: The plot compares the number of epochs for the end-to-end network with each block of QuickNet. Top-right: The plot compares the number of samples in each epoch for the end-to-end network and each block of QuickNet. Bottom-left: Due to the combined effect of fewer epochs and fewer samples per epoch, the total training cost of QuickNet is lower than the end-to-end network. Bottom-right: The cost of training vs. the test accuracy of the network during training. QuickNet have a higher accuracy during training at a lower training cost. }
\label{fig:graph}
\centering
\end{figure*}

\subsection{Performance of QuickNets}
We compare the performance of QuickNets to standard end-to-end networks trained using backpropagation. We also measure the average cost of inference and training for each network. The results are summarized in \ref{sample-table}. Using QuickNets, we are able to speed up inference for MNIST and Fashion-MNIST by 1.11x and 1.14x. In addition the cost of training is reduced by 1.4x and 1.55x respectively. We note that on MNIST and FASHION-MNIST the performance of QuickNet is improved sightly. We hypothesize that this might be due to the reduction in over-thinking \cite{Kaya2019ShallowDeepNU}.

\begin{table*}[t]
\caption{Performance of QuickNets compared to end-to-end training.}
\label{sample-table}
\vskip 0.15in
\begin{center}
\begin{small}
\begin{sc}
\begin{tabular}{llcccc}
\toprule
Data & Method & Accuracy & Parameters (million) & Inference cost & Training cost \\ \midrule
\midrule
\multirow{2}*{MNIST} & QuickNet  & 98.83 $\pm$ 0.08 & 0.141 & 0.124 & 0.57 x $10^{10}$ \\
& End-to-End & 98.33 $\pm$ 0.28 & 0.138 & 0.138 & 0.8 x $10^{10}$ \\
\midrule
\multirow{2}*{FASHION-MNIST} & QuickNet & 87.53 $\pm$ 0.07 & 0.141 & 0.121 & 0.61 x $10^{10}$ \\
& End-to-End & 86.61 $\pm$ 0.72 & 0.138 & 0.138 & 0.95 x $10^{10}$ \\
\midrule
\multirow{2}*{CIFAR-10} & QuickNet & 84.07 $\pm$ 0.24 & 162.45 & 77.72 & 0.86 x $10^{15}$ \\
& End-to-End & 85.41 $\pm$ 0.25 & 28.14 & 172.12 & 1.35 x $10^{15}$ \\
\bottomrule
\end{tabular}
\end{sc}
\end{small}
\end{center}
\vskip -0.1in
\end{table*}

For CIFAR-10, the speed-up during training and testing is more apparent. We are able to increase the inference speed by more than 2x and the training cost by 1.58x. However, due to the block-wise training approach, QuickNet have a reduced accuracy.

\subsection{Reduction in training samples}

In this section, we examine the reduction in the training samples by the QuickNet algorithm and how it enables the reduction in the cost of training. We recorded the number of epoch that each block of QuickNet trained on and the number of training samples in each epoch. As shown in \ref{fig:graph} as the training progresses, it takes fewer epochs to train each block. Additionally, each block is trained on fewer training samples. Due to this, the total cost of training QuickNet is lower than end-to-end training. Note also that Quick-Nets train shallow blocks of network while the end-to-end training would require the error to be propagated through the entire network, thus requiring more compute.

Finally, we also calculate the training cost required for the models to reach a particular accuracy. Due to the block-wise training of our method, the initial blocks can be trained at a very low cost allowing for a better accuracy much earlier when compared to training the entire network end-to-end. Fig. \ref{fig:graph} shows the comparison of training cost vs. accuracy for a VGG-11 network training in an end-to-end fashion and using QuickNet training. We see that using QuickNet training, the net training cost to reach higher accuracies is reduced significantly. This result demonstrates that QuickNets can be deployed faster while training. Each block can be deployed as soon as it is trained since its weights do not change while the deeper blocks are training. 


\subsection{Knowledge distribution across the network}

\begin{figure}[H]
\includegraphics[width=\columnwidth ]{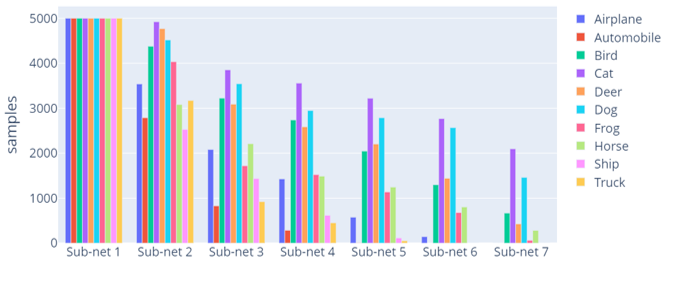}

\caption{ Class distribution of the training samples for each block of QuickNet }
\label{fig:class}
\centering
\end{figure}

To understand the what each block of QuickNet is learning, we further analyze the trained samples by each block. As shown in figure \ref{fig:class}, the first block is trained on the entire training dataset and as the training progresses, the trained samples are removed from the dataset. Interestingly, some classes are learned before others and are removed completely from the training data by the end. Upon further observation, classes of man-made objects like: airplane, automobile, ship and truck seems to be learned faster than classes like bird, cat, dog. Thus, knowledge of man-made objects is automatically assigned the shallower layers of QuickNet while living things like cats, dogs and deer would are learned in the deeper layers.

\subsection{Commitment layers for reliable confidence}

The cross-entropy loss for classification reduces the entropy of the outputs and thus increases the confidence of the network after learning. Confidence is used as a criteria for the decision to exit because it is inversely correlated with the loss. However, it has been demonstrated that for a small subset of inputs, the neural networks have a high confidence for incorrect predictions \cite{Vemuri2020ScoringCI}.

In the QuickNet training algorithm, since the deeper layers are only trained on a subset of the training data, it is imperative to exit at the correct time. Thus, there is a need for a better decision criteria than confidence. However, since main goal of the QuickNet is reduced computation for training and testing, the criteria should not add significant computation for the decision. Therefore, we designed a dual head exit with a classification and a commitment output. The commitment head consist of C hidden neurons and a single output neuron where C is the number of classes.

To demonstrate the efficacy of commitment output, we add it to the state-of-the-art early exit network shallow deep networks (SDN) \cite{Kaya2019ShallowDeepNU}. We then vary the threshold of the confidence for exit for the SDN network and collect the average cost of inference vs accuracy for confidence and confidence + commitment layer.
\begin{figure}[H]
\includegraphics[width=\columnwidth ]{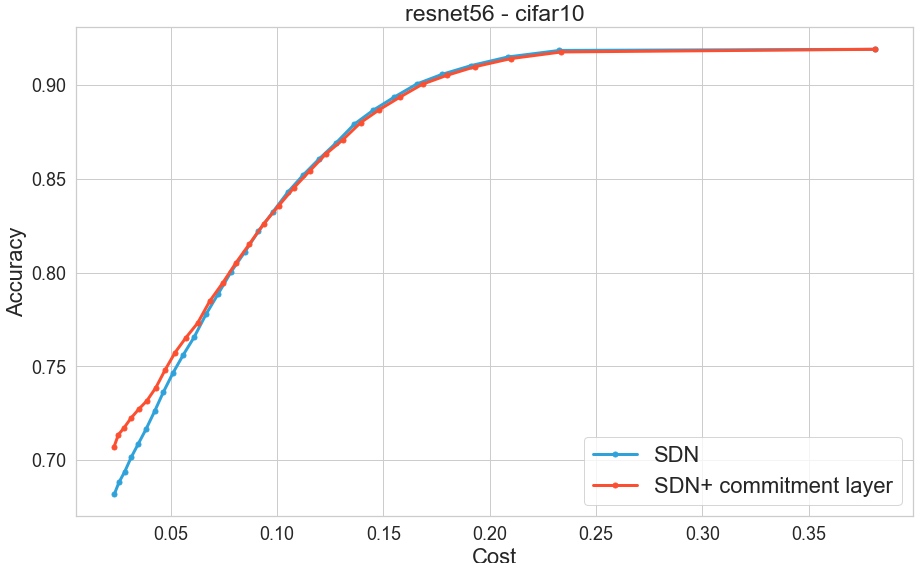}

\caption{ Class distribution of the training samples for each block of QuickNet }
\label{fig:commit}
\centering
\end{figure}

Figure \ref{fig:commit} shows the cost of inference vs. accuracy for confidence and commitment layer on the ResNet-56 CIFAR-10 shallow deep network. We find that commitment layer indeed improves the accuracy for lower computational budgets while achieving a similar performance at higher cost. This demonstrates that commitment layers can be added to existing early-exit approaches for improve their performance. We plan to extend these results to full suit of datasets and architectures that are reported in \citet{Kaya2019ShallowDeepNU}.

\subsection{Scaling to ImageNet}

\begin{table}[t]
\caption{ImageNet Performance on the DenseNet-121 architecture}
\label{img-table}
\vskip 0.15in
\begin{center}
\begin{small}
\begin{sc}
\begin{tabular}{llcccc}
\toprule
Method & Acc. & Params (million) & Inference \\ \midrule
\midrule
QuickNet  & 65.74  & 12.06 & 2279.90 \\
End-to-End & 74.98 & 7.98 & 2882.62 \\

\bottomrule
\end{tabular}
\end{sc}
\end{small}
\end{center}
\vskip -0.1in
\end{table}

In order to demonstrate the scalability of QuickNets, we also demonstrate our approach on the ImageNet dataset \cite{krizhevsky2012imagenet}.  We trained the DenseNet-121 architecture \cite{huang2017densely} in four blocks. We report our results in Table \ref{img-table}. On larger neural networks and datasets, there is a significant decrease in performance but the inference cost is also reduced. In future, we will focus on scaling our method to larger datasets.

\section{Impact and Limitations}

The main hurdle for block-wise training is that it is challenging to match the performance of end-to-end networks for larger architectures. 
Layer-wise training might require architectures that are different than end-to-end architectures. Recently, strides has been made in layer-wise training using the invertible pooling operator. Layer-wise training has demonstrated comparable performance to deep neural networks, however, it also increased the computation cost by an order of magnitude compared to standard architectures like VggNet \cite{Belilovsky2019GreedyLL}. The focus of our work was to reduce the training and inference time simultaneously, therefore, we did not use the architecture in \citet{Belilovsky2019GreedyLL}. 

Block-wise training holds a lot of potential for edge-computing and distributed hardware. Work in this area should focus on better distribution of training data across the sub-networks and architectures that grow in width in addition to the depth. QuickNet constitutes a step towards a truly distributed learning for adaptive training and inference. Further testing of QuickNets on other type of databases, including multi-modal ones, will assist in elevating trust and usage. We envision our approach being applied to a network of connected devices, each containing a part of a network, communicating with each other and each learning a subset of the information.


\section{Conclusion}
We have proposed the Quick-witted neural network training, a novel method for dynamical distribution of training across different layers of the neural network. QuickNets can be implemented on standard deep neural architecture to train them in a block wise fashion for a faster inference and lower training cost by removing trained samples from the training data for deeper layers of the network. We demonstrated our results on fully connected networks and convolutions network architectures. Additionally, we introduced "commitment layers" that can be trained to identify overconfident predictions and thus improving the performance of early-exit neural networks. Finally we postulate that focusing on distributing training and inference across different parts of the network is an important direction for future research and wide-scale deployment of neural networks.

\section*{Acknowledgments}
This material is based upon work supported by the
Defense Advanced Research Projects Agency (DARPA) under Agreement No. HR00112190041. The information contained in this work does not
necessarily reflect the position or the policy of the Government.


\bibliography{paper}  

\begin{thebibliography}{41}
\providecommand{\natexlab}[1]{#1}
\providecommand{\url}[1]{\texttt{#1}}
\expandafter\ifx\csname urlstyle\endcsname\relax
  \providecommand{\doi}[1]{doi: #1}\else
  \providecommand{\doi}{doi: \begingroup \urlstyle{rm}\Url}\fi

\bibitem[Bakhtiarnia et~al.(2021)Bakhtiarnia, Zhang, and
  Iosifidis]{Bakhtiarnia2021ImprovingTA}
A.~Bakhtiarnia, Q.~Zhang, and A.~Iosifidis.
\newblock Improving the accuracy of early exits in multi-exit architectures via
  curriculum learning.
\newblock \emph{ArXiv}, abs/2104.10461, 2021.

\bibitem[Belilovsky et~al.(2019)Belilovsky, Eickenberg, and
  Oyallon]{Belilovsky2019GreedyLL}
E.~Belilovsky, M.~Eickenberg, and E.~Oyallon.
\newblock Greedy layerwise learning can scale to imagenet.
\newblock \emph{ArXiv}, abs/1812.11446, 2019.

\bibitem[Belilovsky et~al.(2020)Belilovsky, Eickenberg, and
  Oyallon]{Belilovsky2020DecoupledGL}
E.~Belilovsky, M.~Eickenberg, and E.~Oyallon.
\newblock Decoupled greedy learning of cnns.
\newblock \emph{ArXiv}, abs/1901.08164, 2020.

\bibitem[Du et~al.(2019)Du, Farrahi, and Niranjan]{Du2019TransferLA}
X.~Du, K.~Farrahi, and M.~Niranjan.
\newblock Transfer learning across human activities using a cascade neural
  network architecture.
\newblock \emph{Proceedings of the 23rd International Symposium on Wearable
  Computers}, 2019.

\bibitem[Enomoto and Eda(2021)]{Enomoto2021LearningTC}
S.~Enomoto and T.~Eda.
\newblock Learning to cascade: Confidence calibration for improving the
  accuracy and computational cost of cascade inference systems.
\newblock \emph{ArXiv}, abs/2104.09286, 2021.

\bibitem[Fahlman and Lebiere(1989)]{Fahlman1989TheCL}
S.~E. Fahlman and C.~Lebiere.
\newblock The cascade-correlation learning architecture.
\newblock In \emph{NIPS}, 1989.

\bibitem[Ghodrati et~al.(2021)Ghodrati, Bejnordi, and
  Habibian]{Ghodrati2021FrameExitCE}
A.~Ghodrati, B.~E. Bejnordi, and A.~Habibian.
\newblock Frameexit: Conditional early exiting for efficient video recognition.
\newblock \emph{ArXiv}, abs/2104.13400, 2021.

\bibitem[Han et~al.(2016)Han, Mao, and Dally]{Han2016DeepCC}
S.~Han, H.~Mao, and W.~J. Dally.
\newblock Deep compression: Compressing deep neural network with pruning,
  trained quantization and huffman coding.
\newblock \emph{arXiv: Computer Vision and Pattern Recognition}, 2016.

\bibitem[Harting et~al.(1992)Harting, Updyke, and van
  Lieshout]{Harting1992CorticotectalPI}
J.~K. Harting, B.~V. Updyke, and D.~P. van Lieshout.
\newblock Corticotectal projections in the cat: Anterograde transport studies
  of twenty‐five cortical areas.
\newblock \emph{Journal of Comparative Neurology}, 324, 1992.

\bibitem[He et~al.(2016)He, Zhang, Ren, and Sun]{he2016deep}
K.~He, X.~Zhang, S.~Ren, and J.~Sun.
\newblock Deep residual learning for image recognition.
\newblock In \emph{Proceedings of the IEEE conference on computer vision and
  pattern recognition}, pages 770--778, 2016.

\bibitem[Huang et~al.(2017)Huang, Liu, Van Der~Maaten, and
  Weinberger]{huang2017densely}
G.~Huang, Z.~Liu, L.~Van Der~Maaten, and K.~Q. Weinberger.
\newblock Densely connected convolutional networks.
\newblock In \emph{Proceedings of the IEEE conference on computer vision and
  pattern recognition}, pages 4700--4708, 2017.

\bibitem[Huang et~al.(2018)Huang, Chen, Li, Wu, Maaten, and
  Weinberger]{Huang2018MultiScaleDN}
G.~Huang, D.~Chen, T.~Li, F.~Wu, L.~V.~D. Maaten, and K.~Q. Weinberger.
\newblock Multi-scale dense networks for resource efficient image
  classification.
\newblock In \emph{ICLR}, 2018.

\bibitem[Kaya et~al.(2019)Kaya, Hong, and Dumitras]{Kaya2019ShallowDeepNU}
Y.~Kaya, S.~Hong, and T.~Dumitras.
\newblock Shallow-deep networks: Understanding and mitigating network
  overthinking.
\newblock In \emph{ICML}, 2019.

\bibitem[Krizhevsky et~al.(2009)Krizhevsky, Hinton,
  et~al.]{krizhevsky2009learning}
A.~Krizhevsky, G.~Hinton, et~al.
\newblock Learning multiple layers of features from tiny images.
\newblock 2009.

\bibitem[Krizhevsky et~al.(2012)Krizhevsky, Sutskever, and
  Hinton]{krizhevsky2012imagenet}
A.~Krizhevsky, I.~Sutskever, and G.~E. Hinton.
\newblock Imagenet classification with deep convolutional neural networks.
\newblock \emph{Advances in neural information processing systems},
  25:\penalty0 1097--1105, 2012.

\bibitem[LeCun et~al.(1998)LeCun, Bottou, Bengio, and
  Haffner]{lecun1998gradient}
Y.~LeCun, L.~Bottou, Y.~Bengio, and P.~Haffner.
\newblock Gradient-based learning applied to document recognition.
\newblock \emph{Proceedings of the IEEE}, 86\penalty0 (11):\penalty0
  2278--2324, 1998.

\bibitem[Lee et~al.(2016)Lee, Gallagher, and Tu]{lee2016generalizing}
C.-Y. Lee, P.~W. Gallagher, and Z.~Tu.
\newblock Generalizing pooling functions in convolutional neural networks:
  Mixed, gated, and tree.
\newblock In \emph{Artificial intelligence and statistics}, pages 464--472.
  PMLR, 2016.

\bibitem[Li et~al.(2019)Li, Zhang, Qi, Yang, and Huang]{Li2019ImprovedTF}
H.~Li, H.~Zhang, X.~Qi, R.~Yang, and G.~Huang.
\newblock Improved techniques for training adaptive deep networks.
\newblock \emph{2019 IEEE/CVF International Conference on Computer Vision
  (ICCV)}, pages 1891--1900, 2019.

\bibitem[Ma et~al.(2020)Ma, Yu, Li, and Wang]{Ma2020WhyLL}
W.~Ma, M.~Yu, K.~Li, and G.~Wang.
\newblock Why layer-wise learning is hard to scale-up and a possible solution
  via accelerated downsampling.
\newblock \emph{2020 IEEE 32nd International Conference on Tools with
  Artificial Intelligence (ICTAI)}, pages 238--243, 2020.

\bibitem[Marquez et~al.(2018)Marquez, Hare, and Niranjan]{Marquez2018DeepCL}
E.~S. Marquez, J.~S. Hare, and M.~Niranjan.
\newblock Deep cascade learning.
\newblock \emph{IEEE Transactions on Neural Networks and Learning Systems},
  29:\penalty0 5475--5485, 2018.

\bibitem[Mostafa et~al.(2018)Mostafa, Ramesh, and
  Cauwenberghs]{Mostafa2018DeepSL}
H.~Mostafa, V.~Ramesh, and G.~Cauwenberghs.
\newblock Deep supervised learning using local errors.
\newblock \emph{Frontiers in Neuroscience}, 12, 2018.

\bibitem[N{\o}kland and Eidnes(2019)]{Nkland2019TrainingNN}
A.~N{\o}kland and L.~Eidnes.
\newblock Training neural networks with local error signals.
\newblock In \emph{ICML}, 2019.

\bibitem[Panda et~al.(2016)Panda, Sengupta, and Roy]{Panda2016ConditionalDL}
P.~Panda, A.~Sengupta, and K.~Roy.
\newblock Conditional deep learning for energy-efficient and enhanced pattern
  recognition.
\newblock \emph{2016 Design, Automation \& Test in Europe Conference \&
  Exhibition (DATE)}, pages 475--480, 2016.

\bibitem[Passalis et~al.(2020)Passalis, Raitoharju, Tefas, and
  Gabbouj]{Passalis2020EfficientAI}
N.~Passalis, J.~Raitoharju, A.~Tefas, and M.~Gabbouj.
\newblock Efficient adaptive inference for deep convolutional neural networks
  using hierarchical early exits.
\newblock \emph{Pattern Recognit.}, 105:\penalty0 107346, 2020.

\bibitem[Scardapane et~al.(2020{\natexlab{a}})Scardapane, Comminiello,
  Scarpiniti, Baccarelli, and Uncini]{Scardapane2020DifferentiableBI}
S.~Scardapane, D.~Comminiello, M.~Scarpiniti, E.~Baccarelli, and A.~Uncini.
\newblock Differentiable branching in deep networks for fast inference.
\newblock \emph{ICASSP 2020 - 2020 IEEE International Conference on Acoustics,
  Speech and Signal Processing (ICASSP)}, pages 4167--4171, 2020{\natexlab{a}}.

\bibitem[Scardapane et~al.(2020{\natexlab{b}})Scardapane, Scarpiniti,
  Baccarelli, and Uncini]{Scardapane2020WhySW}
S.~Scardapane, M.~Scarpiniti, E.~Baccarelli, and A.~Uncini.
\newblock Why should we add early exits to neural networks?
\newblock \emph{Cognitive Computation}, pages 1--13, 2020{\natexlab{b}}.

\bibitem[Shafiee et~al.(2019)Shafiee, Shafiee, and Wong]{Shafiee2019DynamicRT}
M.~S. Shafiee, M.~Shafiee, and A.~Wong.
\newblock Dynamic representations toward efficient inference on deep neural
  networks by decision gates.
\newblock \emph{2019 IEEE/CVF Conference on Computer Vision and Pattern
  Recognition Workshops (CVPRW)}, pages 667--675, 2019.

\bibitem[Simonyan and Zisserman(2014)]{simonyan2014very}
K.~Simonyan and A.~Zisserman.
\newblock Very deep convolutional networks for large-scale image recognition.
\newblock \emph{arXiv preprint arXiv:1409.1556}, 2014.

\bibitem[Skolik et~al.(2021)Skolik, McClean, Mohseni, Smagt, and
  Leib]{Skolik2021LayerwiseLF}
A.~Skolik, J.~McClean, M.~Mohseni, P.~V.~D. Smagt, and M.~Leib.
\newblock Layerwise learning for quantum neural networks.
\newblock \emph{Quantum Machine Intelligence}, 3:\penalty0 1--11, 2021.

\bibitem[Taylor et~al.(2015)Taylor, Hobbs, Burroni, and
  Siegelmann]{Taylor2015TheGL}
P.~Taylor, J.~N. Hobbs, J.~Burroni, and H.~T. Siegelmann.
\newblock The global landscape of cognition: hierarchical aggregation as an
  organizational principle of human cortical networks and functions.
\newblock \emph{Scientific Reports}, 5, 2015.

\bibitem[Teerapittayanon et~al.(2016)Teerapittayanon, McDanel, and
  Kung]{Teerapittayanon2016BranchyNetFI}
S.~Teerapittayanon, B.~McDanel, and H.~T. Kung.
\newblock Branchynet: Fast inference via early exiting from deep neural
  networks.
\newblock \emph{2016 23rd International Conference on Pattern Recognition
  (ICPR)}, pages 2464--2469, 2016.

\bibitem[Veit and Belongie(2018)]{Veit2018ConvolutionalNW}
A.~Veit and S.~J. Belongie.
\newblock Convolutional networks with adaptive inference graphs.
\newblock In \emph{ECCV}, 2018.

\bibitem[Veit et~al.(2016)Veit, Wilber, and Belongie]{Veit2016ResidualNB}
A.~Veit, M.~J. Wilber, and S.~J. Belongie.
\newblock Residual networks behave like ensembles of relatively shallow
  networks.
\newblock In \emph{NIPS}, 2016.

\bibitem[Vemuri(2020)]{Vemuri2020ScoringCI}
N.~Vemuri.
\newblock Scoring confidence in neural networks.
\newblock 2020.

\bibitem[Wang et~al.(2018{\natexlab{a}})Wang, Luo, Crankshaw, Tumanov, and
  Gonzalez]{Wang2018IDKCF}
X.~Wang, Y.~Luo, D.~Crankshaw, A.~Tumanov, and J.~Gonzalez.
\newblock Idk cascades: Fast deep learning by learning not to overthink.
\newblock In \emph{UAI}, 2018{\natexlab{a}}.

\bibitem[Wang et~al.(2018{\natexlab{b}})Wang, Yu, Dou, and
  Gonzalez]{Wang2018SkipNetLD}
X.~Wang, F.~Yu, Z.-Y. Dou, and J.~Gonzalez.
\newblock Skipnet: Learning dynamic routing in convolutional networks.
\newblock In \emph{ECCV}, 2018{\natexlab{b}}.

\bibitem[Wu et~al.(2018)Wu, Nagarajan, Kumar, Rennie, Davis, Grauman, and
  Feris]{Wu2018BlockDropDI}
Z.~Wu, T.~Nagarajan, A.~Kumar, S.~Rennie, L.~Davis, K.~Grauman, and R.~Feris.
\newblock Blockdrop: Dynamic inference paths in residual networks.
\newblock \emph{2018 IEEE/CVF Conference on Computer Vision and Pattern
  Recognition}, pages 8817--8826, 2018.

\bibitem[Xiao et~al.(2017)Xiao, Rasul, and Vollgraf]{xiao2017fashion}
H.~Xiao, K.~Rasul, and R.~Vollgraf.
\newblock Fashion-mnist: a novel image dataset for benchmarking machine
  learning algorithms.
\newblock \emph{arXiv preprint arXiv:1708.07747}, 2017.

\bibitem[Zhang et~al.(2021)Zhang, Do, Doddipatla, Loweimi, Bell, and
  Renals]{Zhang2021TrainYC}
S.~Zhang, C.-T. Do, R.~Doddipatla, E.~Loweimi, P.~Bell, and S.~Renals.
\newblock Train your classifier first: Cascade neural networks training from
  upper layers to lower layers.
\newblock \emph{ArXiv}, abs/2102.04697, 2021.

\bibitem[Zhou et~al.(2020)Zhou, Xu, Ge, McAuley, Xu, and Wei]{Zhou2020BERTLP}
W.~Zhou, C.~Xu, T.~Ge, J.~McAuley, K.~Xu, and F.~Wei.
\newblock Bert loses patience: Fast and robust inference with early exit.
\newblock \emph{ArXiv}, abs/2006.04152, 2020.

\bibitem[Zhou et~al.(2019)Zhou, Bai, Bhattacharyya, and
  Huttunen]{Zhou2019ElasticNN}
Y.~Zhou, Y.~Bai, S.~Bhattacharyya, and H.~Huttunen.
\newblock Elastic neural networks for classification.
\newblock \emph{2019 IEEE International Conference on Artificial Intelligence
  Circuits and Systems (AICAS)}, pages 251--255, 2019.

\end{thebibliography}

\newpage

\appendix
\onecolumn
\section{Training Details}

The networks were train on Pytorch framework on GPU.

For MNIST and FASHION-MNIST, we trained a three-hidden layer neural network in three blocks. We start with the learning rate if 0.001 and reduce it by a factor of 0.5 on plateau with patience 1. The loss was monitored for each epoch and the training was stopped if the minimum loss was not achieved for 3 consecutive epochs. The maximum epochs to be trained were 100. Commitment layers were trained after training the block based on the predictions of the block. Commitment layers were trained in a similar fashion, starting with a learning rate of 0.001 and reducing it on plateau and stopping early. Each early exit consisted of only the output layer for fully connected network.

The end-to-end networks were trained in a similar manner, starting with a learning rate of 0.001 and reducing on plateau with early stopping. 

For the CIFAR-10 network, we used the VGG-11 network with 8 exits. We used the mixed max-average pooling strategy similar to \cite{Kaya2019ShallowDeepNU}. Both the QuickNet and the end-to-end network started with a learning rate of 0.0003 and it was reduced by a factor of 0.5 on plateau with patience 1 with early stopping similar to the fully connected architecture.


\end{document}